\documentclass{article} 
\usepackage{iclr2022_conference,times}

\usepackage{amsmath,amsfonts,bm}

\def\eqref#1{equation~\ref{#1}}

\def\1{\bm{1}}

\DeclareMathAlphabet{\mathsfit}{\encodingdefault}{\sfdefault}{m}{sl}
\SetMathAlphabet{\mathsfit}{bold}{\encodingdefault}{\sfdefault}{bx}{n}

\usepackage{graphicx}
\usepackage{hyperref}
\usepackage{url}
\usepackage{multirow}
\usepackage{arydshln}
\usepackage{xcolor}
\usepackage{tikz}
\usetikzlibrary{tikzmark}
\newlength\savewidth\newcommand\shline{\noalign{\global\savewidth\arrayrulewidth\global\arrayrulewidth 1pt}\hline\noalign{\global\arrayrulewidth\savewidth}}

\title{The impact of spatio-temporal augmentations on self-supervised audiovisual representation learning}

\date{\vspace{-5ex}}
\author{Haider Al-Tahan \\
Department of Computer Science\\
University of Western Ontario\\
London, Ontario, Canada \\
\texttt{haltaha@uwo.ca} \\
\And
Yalda Mohsenzadeh \\
Department of Computer Science\\
University of Western Ontario\\
London, Ontario, Canada \\
\texttt{ymohsenz@uwo.ca} \\
}

\iclrfinalcopy 
\begin{document}

\maketitle

\begin{abstract}
Contrastive learning of auditory and visual perception has been extremely successful when investigated individually. However, there are still major questions on how we could integrate principles learned from both domains to attain effective audiovisual representations. In this paper, we present a contrastive framework to learn audiovisual representations from unlabeled videos. The type and strength of augmentations utilized during self-supervised pre-training play a crucial role for contrastive frameworks to work sufficiently. Hence, we extensively investigate composition of temporal augmentations suitable for learning audiovisual representations; we find lossy spatio-temporal transformations that do not corrupt the temporal coherency of videos are the most effective. Furthermore, we show that the effectiveness of these transformations scales with higher temporal resolution and stronger transformation intensity. Compared to self-supervised models pre-trained on only sampling-based temporal augmentation, self-supervised models pre-trained with our temporal augmentations lead to approximately \textbf{6.5\%} gain on linear classifier performance on AVE dataset. Lastly, we show that despite their simplicity, our proposed transformations work well across self-supervised learning frameworks (SimSiam, MoCoV3, etc), and benchmark audiovisual dataset (AVE).
\end{abstract}

\section{Introduction}

\noindent Visual and auditory perception are the two most utilized human sensory systems in our day-to-day lives. The integration between the two systems allow us to capture rich representations of the environment around us. For instance, in an action recognition task, without sound it can be hard to tell whether a child is babbling, laughing, or coughing using only video frames. However with sound integrated, the problem becomes substantially easier to solve. Applications of learning efficient audiovisual representations can range from audiovisual localization \citep{tian2018audiovisual, arandjelovic2018objects, senocak2018learning} and separation \citep{Zhao_2018_ECCV, gao2018learning, zhao2019sound} to action Recognition \citep{kazakos2019epicfusion, cartas2019seeing, xiao2020audiovisual} and speech recognition \citep{Nagrani17, nagrani2018learnable}.

Recently, contrastive self-supervised learning is at the forefront in learning abstract representations from unlabeled visual or auditory data \citep{he2020moco, chen2020simclr, altahan2020clar, ye2019unsupervised, grill2020byol}. One crucial component that allows these contrastive frameworks to excel at learning representations is the augmentations used during self-supervised pre-training. Hence, extensive augmentation search has been investigated for images \citep{chen2020simclr}, audio \citep{altahan2020clar}, and videos \citep{Feichtenhofer_2021, qian2021spatiotemporal}. Previously proposed video augmentations exploited the temporal dimension of videos, defined as either changing the number of positive clips supplied to the self-supervised loss \citep{Feichtenhofer_2021} or the sampling method of positive clips \citep{qian2021spatiotemporal}. Although, these temporal augmentations were shown to aid in learning better spatio-temporal representations, temporal augmentations that directly change spatial information temporally (spatio-temporal augmentations) while preserving video coherency is yet to be investigated.

In this work, we investigate audiovisual integration in a contrastive self-supervised setting to learn efficient representations. In order to accomplish this, we introduce four major components that are important to nourish the learning of spatio-temporal representations. We:

\begin{itemize}
    \item Demonstrate a simple pipeline to learning efficient audiovisual representations that can be adopted with various existing contrastive frameworks (i.e. SimCLR, MoCo ...).
    \item Introduce four spatio-temporal augmentations that allow models to generalize better to the downstream tasks, compared to control models trained without the proposed spatio-temporal augmentations.
    \item Extensively study hyper-parameters of the spatio-temporal augmentations and verify the effectiveness of the augmentations as we scale the temporal resolutions of videos.
    \item Use the proposed augmentations and investigate temporally aligning augmentations for audiovisual integration.
\end{itemize}

\section{Related Works}

\subsection{Self-Supervised Learning}

\subsubsection{Images}

Self-supervised learning on images has been studied extensively over the last few years, serving as an essential benchmark. Generally self-supervised learning methods exploit the inherit structure of the training data to derive a supervisory signal, called the \textit{pre-text} task. By training on the pre-text task, the aim is to derive effective visual representations from unlabel images, which can be used for various downstream tasks (e.g. classification, segmentation, ...). Some early pre-text tasks leveraged context \citep{pathak2016context}, jigsaw puzzle \citep{noroozi2017jigsaw}, image rotation \citep{gidaris2018rotation}, relative patch spatial location \citep{doersch2016position}, and various other image structural characteristics \citep{zhang2016colorful, zhang2017splitbrain, larsson2017color, bojanowski2017noise}. Numerous pre-text tasks even utilized video frames to learn efficient image representations \citep{wang2015video, vondrick2018tracking, pathak2017learning, gordon2020watching, purushwalkam2020demystifying, jayaraman2016slow}. More recently, \textit{contrastive learning} pre-text tasks has been widely adopted in learning efficient visual representations \citep{he2020moco, chen2020simclr, ye2019unsupervised, grill2020byol, tian2020contrastive, henaff2020dataefficient, oord2019representation, chen2020simsiam}. In short, the learning objective of contrastive learning is to maintain consistent representations between augmented views originating from the same image, while maximizing the representations between views from different images.

\subsubsection{Audio}

Similar to images, auditory data also has been rapidly progressing towards self-supervised auditory representation learning \citep{oord2019representation, wang2021multiformat, altahan2020clar, baevski2020vqwav2vec, baevski2020wav2vec}. Prior works derived efficient auditory representations using videos by predicting whether the visual and audio signals come from the same video \citep{aytar2016soundnet, arandjelovic2017look, arandjelovic2018objects, korbar2018cooperative, owens2018audiovisual, xdc, alayrac2020selfsupervised, patrick2020multimodal}. As discussed earlier, contrastive learning relies heavily on augmented views to construct representations to become less sensitive to sensory-level invariance. \citet{wang2021multiformat} constructed those augmented views for auditory data by contrasting between raw-audio and audio frequency features (e.g. Short-time Fourier transform) using two distinct different models. Alternatively, \citet{altahan2020clar} reduced the reliance on raw-audio by investigating six transformations that specifically tackle auditory data, while only using one model during training. Similar to visual representation learning \citep{chen2020simclr}, some augmentations like pitch shift and fade in/out were shown to result in more efficient representations \citep{altahan2020clar}.

\subsubsection{Audiovisual}

Audiovisual integration using self-supervised learning by exploiting audiovisual correspondence has been extensively explored \citep{arandjelovic2017look, aytar2016soundnet, owens2018audiovisual, owens2016ambient, korbar2018cooperative, hu2019deep, xdc}. The benefit of learning efficient audiovisual representations can aid wide range of tasks beyond video recognition \citep{nagrani2018learnable, arandjelovic2018objects, kazakos2019epicfusion, Zhao_2018_ECCV, cartas2019seeing, zhao2019sound}. \citet{qian2021spatiotemporal} investigated the spatio-temporal component of video frames using contrastive learning for learning video representations. \citet{qian2021spatiotemporal} found that maintaining temporal consistency in-regard to spatial augmentation is crucial for better representations. Furthermore, they found that the temporal sampling strategy for the contrasted positive clips is also important in learning efficient representations. Alternatively, \citet{Feichtenhofer_2021} defined temporal augmentations as the number of clips sampled at different temporal locations as positive samples and found that by increasing the number of clips ($\geq 2$), we can obtain better representations.

In this work, we adopt findings from mono-domain contrastive learning frameworks on audio and video domains. In particular, for the video domain, we utilize the sampling strategy with a monotonically decreasing distribution \citep{qian2021spatiotemporal}. Furthermore, we maintain consistent spatial augmentations across frames within each clip. For the audio domain, we utilized the auditory augmentations that affect the frequency structure of the signal rather than the temporal component of a signal such as pitch shift \citep{altahan2020clar}. Lastly, to nourish better audiovisual representations, we introduce four spatio-temporal augmentations and extensively study the effectiveness of these augmentations on generalizability to downstream tasks.

\section{Methods}

\subsection{Contrastive Learning Frameworks}

Contrastive Learning frameworks generally aims to maximize similarity between representations of the same samples that are augmented differently $\tilde{x}$ (positive samples) and minimize similarity between representations of different samples (negative samples). In this paper, our positive samples are the set of clips extracted from each video and our contrastive loss in principle follows the InfoNCE objective \citep{oord2019representation, chen2020simclr}: 

\begin{equation}
    \mathcal{L}_{i,j} = -\log\frac{\exp\left(\text{sim}\left(\mathbf{z}_{i}, \mathbf{z}_{j}\right)/\tau\right)}{\sum^{2N}_{k=1}\mathbf{1}_{[k\neq{i}]}\exp\left(\text{sim}\left(\mathbf{z}_{i}, \mathbf{z}_{k}\right)/\tau\right)}
    \label{simclrloss}
\end{equation}

\noindent where $\mathcal{L} = \sum^{N}_{i,j} \mathcal{L}_{i,j}$, $\mathbf{1}_{[k\neq{i}]} \in 0, 1$ is an indicator function evaluating to $1$ iff $k\neq{i}$, $\tau$ denotes a temperature parameter, and $\mathbf{z}$ denotes $\ell_2$ normalized encoded representations of a given augmented clip. $N$ is the number of training samples within a mini-batch,  $\left(i, j\right)$ are positive clips from each video. The loss is computed across all positive clips, in a mini-batch. $\text{sim}\left(\mathbf{u}, \mathbf{v}\right) = \mathbf{u}^{T}\mathbf{v}/||\mathbf{u}|| ||\mathbf{v}||$ denotes the cosine similarity between two vectors $\mathbf{u}$ and $\mathbf{v}$. 

In this paper, we investigate multitude of frameworks for audiovisual representations learning that utilizes a variant of the InfoNCE objective:

\begin{enumerate}
    \item \textbf{SimCLR} \citep{chen2020simclr} uses the same objective as Eq.\ref{simclrloss}, where the representations of different clips within a mini-batch are treated as the negative samples. SimCLR adopt the encoder to extract the representations of different clips to compute the contrastive loss with the gradient of both views flowing through the encoder. Hence, this method relies heavily on negative samples to prevent collapsing.
    \item \textbf{MoCo} \citep{chen2021mocov3} replaces the second identical encoder with a momentum encoder, $\theta_m$. The momentum encoder parameters are a moving average of the encoder $\theta$ and updated at each step: $\theta_m \leftarrow m\theta_m + (1-m)\theta$ where $m \in (0,1)$ is the hyper-parameter that dictates the degree of change. There is no gradient flowing through the momentum encoder. Our implementation follows \citet{chen2021mocov3} design choices rather than \citet{chen2020mocov2}; this means that we replace shuffling batch normalization (BN) with sync BN, the projection head is a 3-layer MLP, and the prediction head is a 2-layer MLP. The prediction head is stacked on top of the projection head, however, the prediction head does not stack on top of momentum encoder representations. Lastly, we preserve the memory queue, as we utilize relatively small batch size.
    \item \textbf{BYOL} \citep{grill2020byol} can be viewed as a variant of MoCo that does not use negative samples. Hence, the loss function consists of only the numerator part of Eq. \ref{simclrloss}.
    \item \textbf{SimSiam} \citep{chen2020simsiam} similar to SimCLR, this method uses identical second encoder. However, SimSiam does not use negative samples and the gradient only flow through one of the encoders. SimSiam can be thought of as BYOL without the momentum encoder.
    
\end{enumerate}

\subsection{Audiovisual Encoder}

For each clip, we encode the video frames and audio spectrograms using a ResNet \citep{he2015resnet} variant. Unless mentioned otherwise, we utilized ResNet-18 for both streams. For the video encoder, we follow the design of SlowFast network with the modifications proposed in CVRL \citep{feichtenhofer2019slowfast, qian2021spatiotemporal}. For the audio encoder, we followed the design of \citep{altahan2020clar, kazakos2021slowfast}, however due to memory restrains we apply max-pooling to the temporal dimension, contrary to the implementation proposed by \citet{kazakos2021slowfast}. All models were trained from random initialization with 4 and 8 NVIDIA v100 Tesla GPUs.


\subsection{Data Augmentations}

\begin{figure}[t]
\centering
\includegraphics[width=0.7\textwidth]{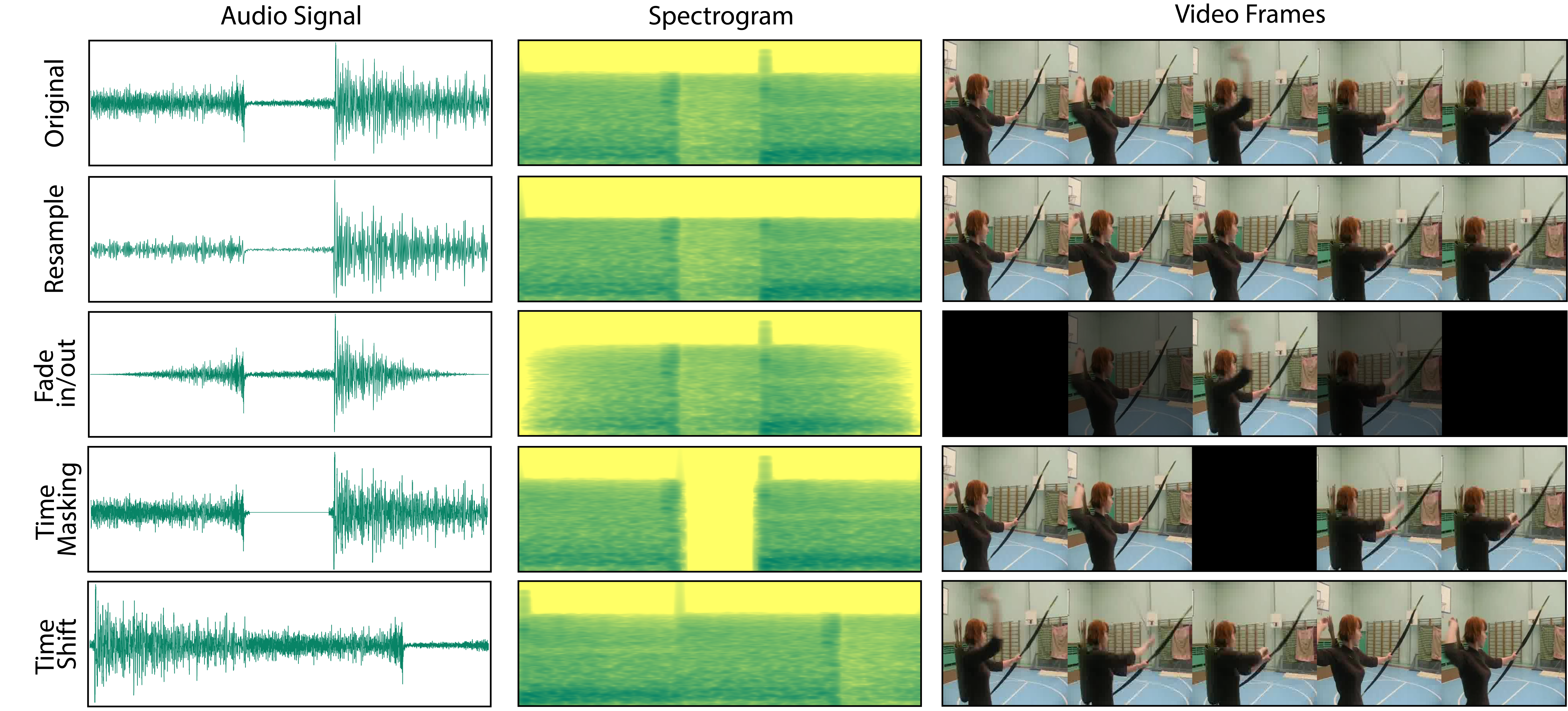}
\caption{Temporal Augmentations. Each row demonstrate the effect of an augmentation on the raw waveform, mel-spectrogram, and video frames, respectively. The first row, represent the original video with its respective audio without any transformation applied.}
\label{fig:augmentations}
\end{figure}


The choice of augmentations are crucial component of contrastive learning to construct view invariant in which makes the model learn efficient representations and become less sensitive to variations in the input data \citep{chen2020simclr, altahan2020clar, gidaris2018rotation, noroozi2017jigsaw}. For images, \citet{chen2020simclr} have shown that the combination of random cropping followed by resize back to the original size, random horizontal flip, random color distortions, and random Gaussian blur is crucial to achieve a good performance. For sounds, \citet{altahan2020clar} have shown that the combination of frequency and temporal guided augmentations yield the best performance (i.e. pitch shifting and time masking). We incorporate augmentations from both domains in our framework, however for transformations that operate on the temporal axis, we apply them for both auditory and visual data (see Section \ref{sec_aug}). For training dataset, we utilize AVE dataset \citep{tian2018audio} because audios and videos correspond, in the sense that the sound source is always visually evident within the video clip.

\subsubsection{Temporal Augmentations}\label{sec_aug}

Figure \ref{fig:augmentations} shows the effect of temporal augmentations on raw waveform, mel-spectrogram, and video frames, respectively. For each augmentation we defined parameter $\alpha \in [0,1]$ which controls the maximum intensity of the augmentations across the temporal dimension; where $0$ in-tales that none of the video is augmented and $1$ means that the augmentation affect all the video in the temporal dimension. In the current section, we will describe the specifics of each augmentation: 

\begin{enumerate}
    \item \textbf{Fade in/out (FD)}: Gradually increases/decreases the intensity of the audio signal or video frames starting from the beginning and end of the video to the middle of the video. The degree of the fade is either linear, logarithmic, exponential, quarter sine, or half sine (applied with uniform probability distribution). The size of the fade for either side of the video is another random parameter applied using a uniform probability distribution (each side of the fade is independent). For instance, $\alpha = 0.5$ for one side means that $1/4$ of the total video clip is used for the fade effect. 
    \item \textbf{Time Masking (TM)}: Randomly masks a small segment of the audio signal or video frames with normal noise or a constant value. We randomly selected the location of the masked segment and the size of the segment from a uniform distribution. $\alpha$ controls for the maximum size of the masked segment.
    \item \textbf{Time Shift (TS)}: randomly shifts the audio signal or video frames forwards or backwards temporally. Samples/frames that roll beyond the last position are re-introduced at the first position (rollover). The degree and direction of the shifts were randomly selected for each audio. The maximum degree that could be shifted was based on $\alpha$, while, the minimum was when no shift applied.
    \item \textbf{Resample (RE)}: Randomly apply temporal down-sample, followed by an up-sample back to the original shape. The RE transformation results in lower resolution audio and video signals with repetitive samples. For instance, $\alpha = 0.5$ for an audio signal with 44.1 kHz sampling rate would be down-sampled to 22 kHz and up-sampled back to 44.1 kHz. Further for video frames with 8 fps, $\alpha = 0.5$ would down-sample the video to 4 fps and up-sample back 8 fps.
\end{enumerate}

\subsubsection{Temporal Augmentation Alignment}

Augmentations that maintain the spatio-temporal coherency within videos have been shown to yield better representations \citep{qian2021spatiotemporal}. We incorporate an analogous paradigm by aligning augmentations temporally across audio and video streams. For instance, with fade in/out augmentation, we align the location of the fade (beginning or end) and strength of the augmentation by making them consistent across the two streams. Similar to the fade in/out augmentation, time masking and time shift follow the same standard; where the strength and the temporal location of the augmentation is made to be consistent across streams. Lastly, for the resample augmentation, we simply maintain consistency just for the strength ($\alpha$) of the augmentation.

\begin{figure}[t]
\centering
\includegraphics[width=0.75\textwidth]{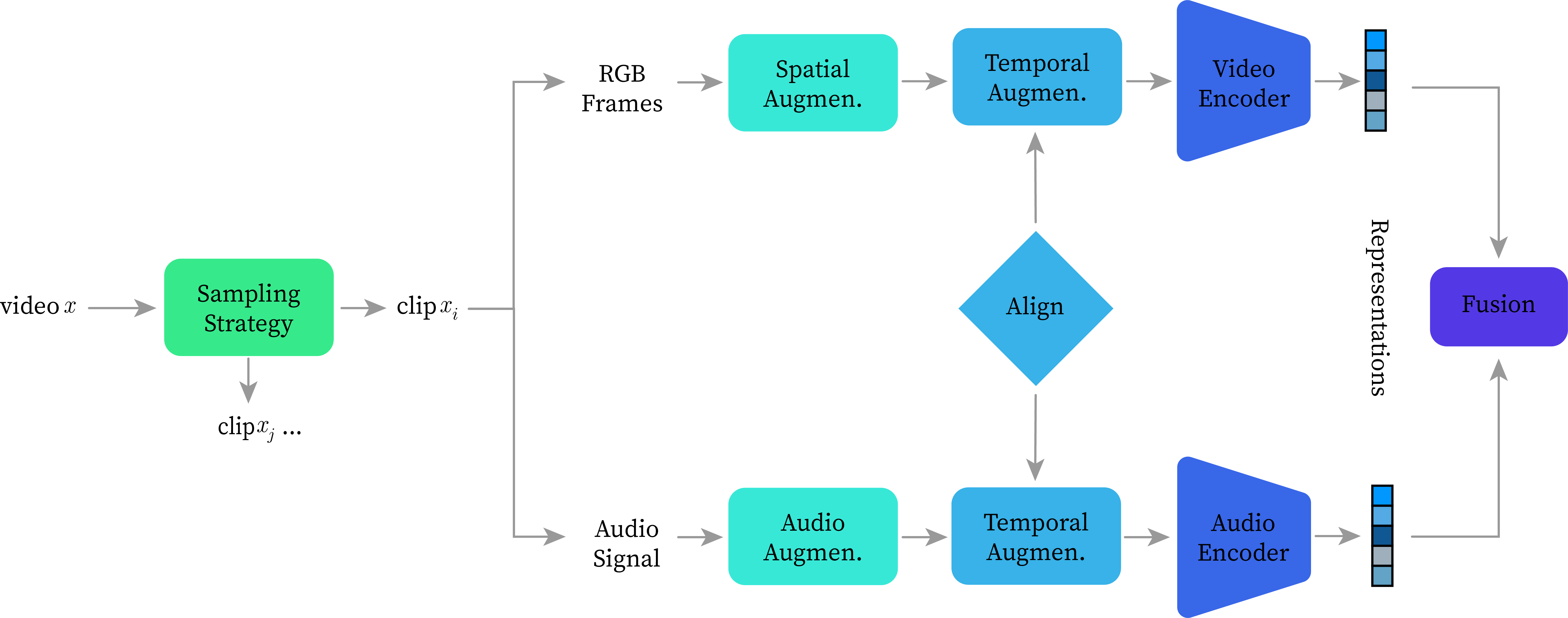} 
\caption{Illustration of self-supervised pre-training pipeline. Given a video $x$, we sample 2 clips using monotonically decreasing probability distribution, followed by augmenting each stream with domain specific augmentations and temporal augmentations. 
}
\label{fig:method}
\end{figure}

\subsection{Evaluation}

To evaluate the efficiency of learned representations in our experiments, we follow common practice in self-supervised learning literature \citep{chen2020simclr, he2020moco}. Specifically, after training the encoder on the contrastive learning pre-text task, we fix the weights in the encoder and train a linear classifier on top of it to evaluate the efficiency of the encoder learned representations. During the training of the linear classifier, we only train using domain specific augmentations, without the temporal augmentations. This measure was done to maintain consistent pipeline when comparing encoders trained using different temporal augmentations (i.e. \ref{results_temporal}). The linear classifier training uses base $lr$ = 30 with a cosine decay scheduler for 100 epochs, weight decay = 0, momentum = 0.9, batch size = 256 and SGD optimizer, following \citet{chen2020simsiam, he2020moco}. During testing, we follow common practice of uniformly sampling 10 clips from each video with a 3-crop evaluation \citep{feichtenhofer2019slowfast, wang2018nonlocal} where each clip has spatial shape of $256 \times 256$. The final prediction is the averaged softmax scores of all clips.

\subsection{Implementation Detail}

Following \citet{chen2020simsiam}, all experiments were trained using SGD optimizer with a learning rate of $lr \times$ Batch Size / 256 ($lr = 0.1$), momentum of $0.9$, and weight decay of $10^{-4}$. Weight decay were not applied to bias parameters and batch normalization layers. Furthermore, batch normalization layers were synchronized across devices. For learning rate, we used linear warm-up for the first 10 epochs \citep{goyal2018accurate}, and cosine decay schedule without restarts \citep{loshchilov2017sgdr}. Unless specified, we used batch size of $128$ for all experiments. Consistent with \citep{qian2021spatiotemporal}, each clip is 1.28 seconds long and sampled the two positive clips using the same strategy. That is, given an input video $T$, we draw a time interval $t$ from a monotonically decreasing distribution over $[0, T]$. Following that, we uniformly sample the first clip interval from $[0, T-t]$, while the second clip is delayed by $t$ after the first.

\section{Results}

\subsection{Auditory Augmentations} \label{results_auditory}

Auditory data can be augmented in both temporal and frequency domain \citep{altahan2020clar}. For the audio stream, we initially adopt frequency augmentations followed by temporal augmentations. \citet{altahan2020clar} have shown colored noise and pitch shifting as viable options. Table\ref{tab:audio_aug} investigates the effect of different auditory augmentations and the location of applying the auditory augmentations relative to the temporal augmentations on linear classifier top-1/5 accuracy after being pre-trained (with contrastive loss) on AVE \citep{tian2018audio} dataset for 500 epochs. Results suggest that pitch shifting on average shows \textbf{+5.9\%} higher performance compared to colored noise. This result is consistent with \citep{altahan2020clar}. Furthermore, we observe that placing the auditory augmentations either before/after the spatio-temporal augmentations can yield $\sim$\textbf{1\%} difference in performance. This can also depend on the auditory augmentation applied, for instance, pitch shifting benefit when placed before the spatio-temporal augmentations, while colored noise benefit when placed after. To stay consistent, moving forward we will only use pitch shifting before the spatio-temporal augmentations.

\begin{table}[th]
\caption{The impact of type and location of audio augmentations applied before temporal augmentations. RE was applied for pre-training all models. Top-1/5 accuracy of linear classifiers pre-trained on AVE dataset for 500 epochs.} \vspace{4pt}
    \label{tab:audio_aug}
    \centering
    \def\arraystretch{1.25}
    \begin{tabular}{l@{\hskip .25in}|c@{\hskip .25in}c@{\hskip .25in}c@{\hskip .25in}c}
    \shline
    \multirow{4}{*}{\begin{tabular}[c]{@{}c@{}}Audio \\Augmentation\end{tabular}} & \multicolumn{4}{c}{Location}  \\        & \multicolumn{2}{c}{Before}        & \multicolumn{2}{c}{After}         \\ \cline{2-5} 
    & \multicolumn{2}{c}{Accuracy (\%)} & \multicolumn{2}{c}{Accuracy (\%)} \\   & top-1            & top-5          & top-1           & top-5           \\ \shline
    Colored Noise   & 58.48           & 90.55           & 59.95  & 92.04           \\
    Pitch Shift   & \textbf{65.42}  & \textbf{93.28}  & 64.18  & 91.29 \\ \shline
    \end{tabular}
    
\end{table}

\subsection{Temporal Augmentations} \label{results_temporal}

Most recent contrastive learning frameworks relies heavily on augmentations to derive the contrastive loss. While previous video self-supervised frameworks investigated the effect of temporal sampling of clips on learnt representations \citep{qian2021spatiotemporal}, a wider range of temporal augmentations that can directly effect the learned representations has not been explored. In this section, we vary the strength, sequence, and across domain application of the proposed spatio-temporal augmentations to explore their effect on predictive performance:

\subsubsection{Strength}

Before exploring multiple spatio-temporal augmentations in sequence, we first investigate the effect of changing the strength of each temporal augmentation during the training on the contrastive pre-task. Table \ref{tab:temp_aug_alpha} shows the effect of varying $\alpha$ for each of the spatio-temporal augmentations. As described in Section \ref{sec_aug}, $\alpha$ controls the maximum intensity of the augmentations across the temporal dimension. Similar to prior work using spatial augmentations \citep{chen2020simclr, Feichtenhofer_2021}, we observe that stronger augmentations (higher $\alpha$) generally increase performance of the linear classifier. With the exception being Time Shift, where performance does not significantly change across different $\alpha$. The best performing augmentation is Resample with \textbf{+5.0\%} higher performance than the second best performing temporal augmentation. Furthermore, Resample shows the most performance difference when changing $\alpha$, with \textbf{+6.5\%} difference going from low to high $\alpha$.

\begin{table}[th]
\caption{The impact of $\alpha$ variable for each of the augmentations. Top-1 accuracy of linear classifiers pre-trained on AVE dataset for 500 epochs.} \vspace{4pt}
    \label{tab:temp_aug_alpha}
    \centering
    \def\arraystretch{1.25}
    \begin{tabular}{l@{\hskip .25in}|c@{\hskip .25in}c@{\hskip .25in}c}
    \shline
    \multirow{2}{*}{Augmentation}  & \multicolumn{3}{c}{Top-1 Accuracy (\%)} \\ 
                  & $\alpha=0.25$     & $\alpha=0.50$    & $\alpha=0.75$    \\ \shline
    Resample      &  58.96           & 62.69     & \textbf{65.42}     \\ 
    Fade in/out   &  49.75           & \textbf{58.71}     & 57.21             \\ 
    Time Masking  &  54.73           & 57.71     & \textbf{60.45}     \\ 
    Time Shifting &  \textbf{57.96}           & 56.72     & 57.71    \\ \shline
    \end{tabular}
    
\end{table}

\subsubsection{Sequence}\label{sec:seq}

\begin{figure}[th]
\centering
\includegraphics[width=0.45\columnwidth]{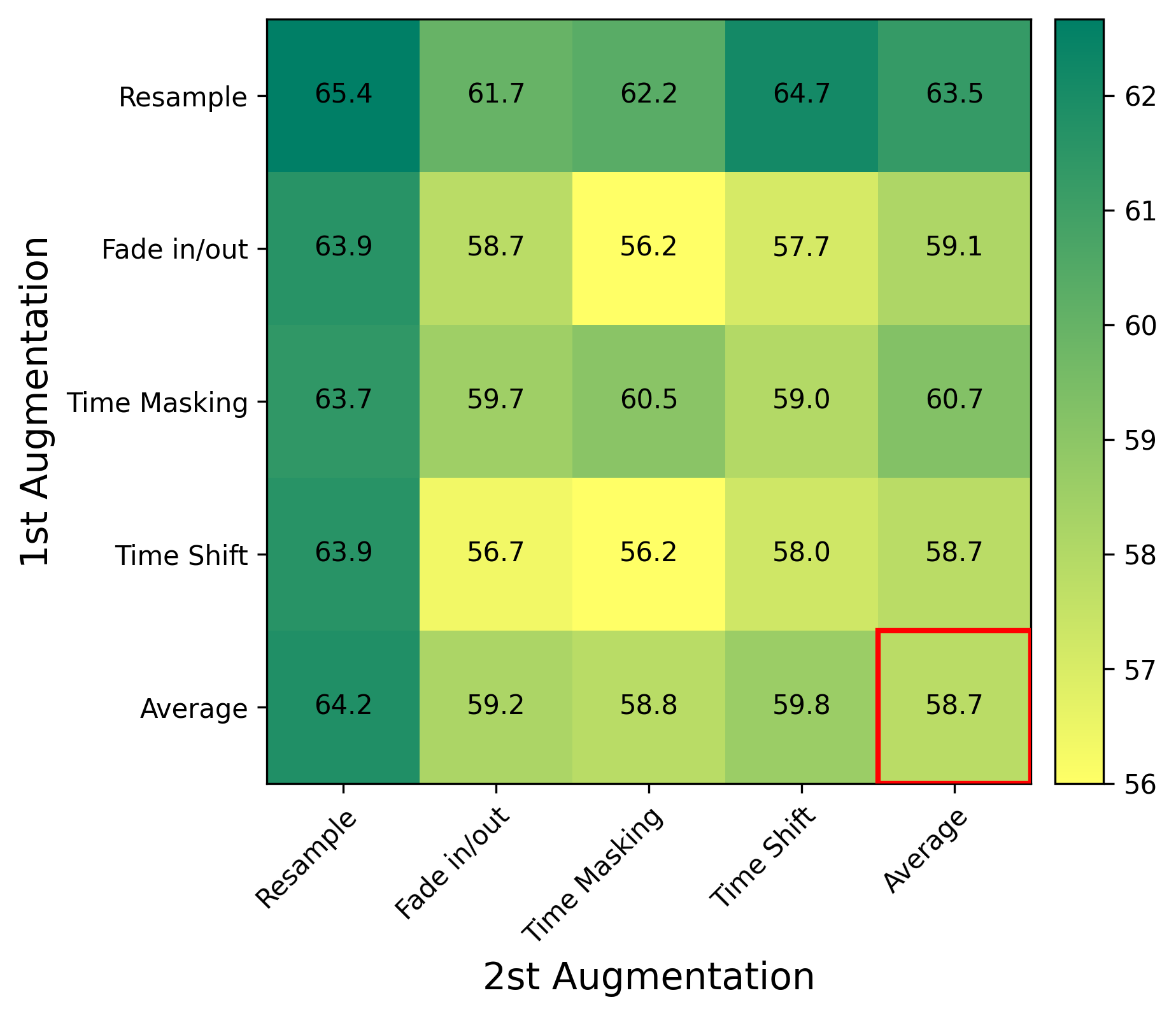} 
\caption{Top-1 accuracy of linear classifiers on representations pre-trained with contrastive loss on AVE dataset using one or pair of spatio-temporal augmentations for 500 epochs. The bottom-right element is a model trained with no spatio-temporal augmentation. The last row/column represents the average of each row/column, respectively. Diagonal of the matrix demonstrates the result when only one corresponding spatio-temporal augmentation applied during pre-training.}
\label{fig:seq}
\end{figure}

To systematically investigate the impact of temporal augmentations and their sequential ordering, we explore a composition of the proposed augmentations during pre-training on the contrastive task. Figure \ref{fig:seq} shows linear classifier top-1 performance on AVE dataset. The diagonal line represents the performance of single augmentation, while other entries represent the performance of paired augmentations. Each row indicates the first augmentation and each column shows the second augmentation applied sequentially. The last column depicts the average when the augmentation was applied first, while the last row shows the average performance over the corresponding augmentations when they were applied the second. The bottom right element serve as a control where no spatio-temporal augmentation was applied (\textbf{58.7\%}). We observe that Resample plays a critical role in learning good representation. As just by including the Resample transformations during contrastive pre-training, we at least attain \textbf{61.7\%}, which outperforms all models that are pre-trained on other spatio-temporal augmentations.

\subsubsection{Across Domain Application}

To investigate the impact of these spatio-temporal augmentations on the learnt representations of multiple domains, we restrain the temporal augmentations to one domain and explore across domains temporal alignment. In Table \ref{tab:temp_aug_loca}, we select the two top competing augmentations from Section \ref{sec:seq}: Resample only (\textbf{65.4\%}) and Resample + Time Shift (\textbf{64.7\%}). Without spatio-temporal augmentations, the linear classifier top-1 accuracy is \textbf{58.7\%}. We observe that on average compared to no spatio-temporal augmentations, applying these augmentations only to the audio stream produces better representations compared to video stream (\textbf{5.23\%} vs \textbf{2.12\%}; respectively). We further tested whether aligning the intensity ($\alpha$) and location (does not apply for Resample) of the augmentations temporally would result in better representations. However, we observe that on average not aligning the augmentations temporally produces better representations (\textbf{+2.37\%} improvement).

\begin{table}[t]
\caption{Investigating the impact of location and alignment of the spatio-temporal augmentations. Top-1/5  accuracy  of  linear  classifiers on representations  pre-trained with contrastive loss using the two best performing spatio-temporal augmentations on AVE dataset for 500 epochs.}
\vspace{4pt}
    \label{tab:temp_aug_loca}
    \centering
    \def\arraystretch{1.25}
    \begin{tabular}{l|cccc}
\shline
\multirow{4}{*}{\begin{tabular}[c]{@{}c@{}}Augmentation \\Location\end{tabular}} & \multicolumn{4}{c}{Augmentation}  \\        & \multicolumn{2}{c}{RE}        & \multicolumn{2}{c}{RE + TS}         \\ \cline{2-5} 
& \multicolumn{2}{c}{Accuracy (\%)} & \multicolumn{2}{c}{Accuracy (\%)} \\   & top-1            & top-5          & top-1           & top-5           \\ \shline
No Aug.                 & 58.70           & 90.55           & 58.70  & 90.55           \\
Audio only              & 63.68           & 92.29           & 64.18  & 93.03           \\
Video only              & 59.95           & 91.54           & 61.69  & 91.04           \\
Both (+ alignment)      & 63.43           & 91.79           & 61.44  & 93.28           \\
Both (+ no alignment)   & \textbf{65.42}  & \textbf{93.28}  & 64.18  & 91.29 \\ \shline
\end{tabular}
\end{table}

We verify the effectiveness of the proposed spatio-temporal augmentations by varying the temporal resolution of both video and audio streams independently. Table \ref{tab:temp_aug_resolution} demonstrates that we are able to better capitalize on higher video and/or audio temporal resolution when using spatio-temporal augmentations compared to no spatio-temporal augmentation control models. For video frames, when varying the number of frames we observe higher performance for spatio-temporal augmentations compared to control \textbf{+7.71\%}. Furthermore, we observe that by only increasing the number of frames from 4 to 32 during training, the gap in performance between control and spatio-temporal augmentation models increases by \textbf{+3.98\%}. Alternatively, for audio temporal resolution we vary two parameters: spectrogram hop size and sampling rate (kHz). Although spectrogram hop size increases the amount of resolution of spectrograms across the temporal dimension, we find no benefit in increasing the spectrogram resolution. Because the gap in performance between control and spatio-temporal augmentation degrades when moving from 256 to 32 hop size (\textbf{-3.73\%}). Similar to video frames, we observe that by increasing the audio sampling rate the gap in performance between control and spatio-temporal augmentation models increases when moving from 5.5 kHz to 44.1 kHz (\textbf{+2.48\%}). In this paper, when not specified, the default number of frames is 8, spectrogram hop size is 128, and audio sampling rate is 44.1kHz.

\begin{table*}[th]
\caption{Investigating the impact of temporal resolution of audio and video streams. Temporal resolution for video frames is adjusted through number of frames given to the model during pre-training on AVE dataset. For the audio signal, we adjusted the spectrogram hop size and sampling rate (kHz) during pre-training. Each section shows the effect of increasing temporal resolution with and without spatio-temporal augmentations on performance.} \vspace{4pt}
    \label{tab:temp_aug_resolution}
    \centering
    \def\arraystretch{1.3}
\begin{tabular}{lcccccc}
\shline
\multicolumn{1}{l}{Temporal Augmentation} & \multicolumn{4}{c}{Temporal Resolution} \\ 
& \textit{lowest}\tikzmark{a} & & & \tikzmark{b}\textit{highest} \\
Number of Frames & 4                    & 8                  & 16     & \multicolumn{1}{c:}{32} & $\bar{x}$ &    \\ \shline
No Augmentation & 57.71                & 58.71              & 58.46  & \multicolumn{1}{c:}{54.73} & 57.40 &\multirow{2}{*}{\textbf{+7.71}}\\
Resample        & 63.93                & 65.42              & 66.17  & \multicolumn{1}{c:}{64.93} & 65.11 &\\ \hdashline
$\Delta(x)$      & 6.22   & 6.71 & 7.71 & \multicolumn{1}{c:}{10.2} &  \\
 & \multicolumn{4}{c:}{\hfill\null \textbf{+0.49} \hfill\null \textbf{+1.00} \hfill\null \textbf{+2.49}} \hfill\null& \\ \shline
Spectrogram Hop Size & 256                    & 128               &  64       & \multicolumn{1}{c:}{32} & $\bar{x}$ &  \\ \shline
No Augmentation & 55.97                & 58.71              &   54.48 & \multicolumn{1}{c:}{54.48} & 55.91 &\multirow{2}{*}{\textbf{+6.84}}\\
Resample        &  64.18                & 65.42              &  62.44 & \multicolumn{1}{c:}{58.96} & 62.75\\ \hdashline
$\Delta(x)$      & 8.21    & 6.71 & 7.96 & \multicolumn{1}{c:}{4.48} & \\ 
& \multicolumn{4}{c:}{\hfill\null $-$1.50 \hfill\null \textbf{+1.25} \hfill\null $-$3.48} \hfill\null& \\ \shline
Audio Sampling Rate (kHz) & 5.5                    & 11               & 22 & \multicolumn{1}{c:}{44.1} & $\bar{x}$ &   \\ \shline
No Augmentation & 53.48                &        54.98       & 60.20   & \multicolumn{1}{c:}{58.71} & 56.84&\multirow{2}{*}{\textbf{+3.92}}\\
Resample        &  57.71                &       57.21        &  62.69 & \multicolumn{1}{c:}{65.42} & 60.75\\ \hdashline
$\Delta(x)$      & 4.23    & 2.23 & 2.49 & \multicolumn{1}{c:}{6.71} & \\ 
& \multicolumn{4}{c:}{\hfill\null $-$2.00 \hfill\null \textbf{+0.26} \hfill\null \textbf{+4.22}} \hfill\null& \\ \shline
\end{tabular}
    \begin{tikzpicture}[overlay, remember picture, shorten >=5pt, shorten <=5pt, transform canvas={yshift=.25\baselineskip}]
        \draw [->, thick] ({pic cs:a}) [left] to ({pic cs:b});
    \end{tikzpicture}
\end{table*}

\subsection{Contrastive Learning Frameworks}

Over the past few years, abundant number of contrastive frameworks has been proposed \citep{he2020moco, chen2020simclr, grill2020byol, chen2020simsiam}. In this section, we explore different frameworks and investigate their performance on learning efficient audiovisual representations. Table \ref{tab:contrast_method} demonstrates the linear classifier top-1 accuracy of various contrastive frameworks on AVE dataset. We observe that frameworks (i.e. SimCLR and MoCoV3) that incorporate negative samples in their loss, generally require less training to reach competitive representations compared to frameworks (i.e. BYOL and SimSiam) that rely only on positive samples. \citet{Feichtenhofer_2021} showed similar trend with BYOL and MoCoV2 on K400 and UCF101 datasets trained only on video frames. Although SimCLR and MoCoV3 outperform other methods when pre-trained with $\leq 400$ epochs, BYOL and SimSiam produce better representations when trained longer (i.e. 800 epochs). Lastly, SimSiam outperforms all methods with \textbf{71.64\%} accuracy. 

\begin{table*}[th]
    \caption{Investigating the impact of number of self-supervised pre-training epochs and the type of self-supervised methods used. The same augmentations were used for all the models.} \vspace{4pt}
    \label{tab:contrast_method} 
    \centering
    \def\arraystretch{1.3}
    \begin{tabular}{l@{\hskip .3in}|c@{\hskip .3in}c@{\hskip .3in}c@{\hskip .3in}c}
    \shline
    \multirow{3}{*}{\begin{tabular}[c]{@{}c@{}}Pre-training\\Method\end{tabular}} & \multicolumn{4}{c}{Top-1 Accuracy (\%)}   \\ \cline{2-5}
    & \multicolumn{4}{c}{Pre-training Duration} \\
     & 100 & 200 & 400 & 800 \\ \shline
    \color{gray} Supervised & 66.17  & 70.90  & 71.89 & 74.88  \\ \color{black}
    MoCoV3 & 58.21  & 59.95  & 62.44 & 66.92  \\
    SimCLR & 59.45  & 63.18  & 66.17 & 68.66 \\
    BYOL & 53.73  & 55.72  & 63.43 & 68.66 \\
    SimSiam & 49.00  & 51.99 & 63.18 & \textbf{71.64}  \\ \shline
    \end{tabular}
\end{table*}

\section{Conclusion}

In this paper, with extensive and comprehensive experiments on various design choices and audiovisual augmentations, we proposed an effective new pipeline for audiovisual contrastive learning. 
Together, our results depict a promising path towards automated audiovisual integration to learn efficient audiovisual representations from unlabeled videos.

\bibliography{iclr2022_conference}
\bibliographystyle{iclr2022_conference}


\end{document}